\pdfoutput=1

\documentclass[11pt]{article}

\usepackage[]{EMNLP2023}

\usepackage{times}
\usepackage{latexsym}

\usepackage[T1]{fontenc}

\usepackage[utf8]{inputenc}

\usepackage{microtype}

\usepackage{inconsolata}
\usepackage{float}
\usepackage{amsfonts}
\usepackage{pdfpages}

\usepackage{color}
\usepackage{algorithm}
\usepackage{algorithmic}
\usepackage{amsmath}
\usepackage{amsthm}
\usepackage{enumitem}
\usepackage{bm}
\usepackage{subfigure} 
\usepackage{nicefrac}
\usepackage{tabularx}
\usepackage{url}
\usepackage{booktabs}
\usepackage{multirow}
\usepackage{graphicx}
\usepackage{amssymb}
\usepackage{bbding}

\graphicspath{ {./images/} }

\newcolumntype{L}[1]{>{\raggedright\let\newline\\\arraybackslash\hspace{0pt}}m{#1}}
\newcolumntype{C}[1]{>{\centering\let\newline  \\\arraybackslash\hspace{0pt}}m{#1}}
\newcolumntype{R}[1]{>{\raggedleft\let\newline \\\arraybackslash\hspace{0pt}}m{#1}}

\title{Loss-aware Curriculum Learning for Heterogeneous Graph Neural Networks}

\author{Zhen Hao Wong$^a$ \quad Hansi Yang$^b$ \quad Xiaoyi Fu$^c$ \quad Quanming Yao$^a$ \\
	$^a$  Department of Electronic Engineering, Tsinghua University, \\
 $^b$  Department of Computer Science and Engineering, Hong Kong University of Science and Technology\\
  $^c$  E. Copi (ELN Copilot) Team, KongFoo Technology Co., Ltd.\\
        \texttt{\small huang-zh20@mails.tsinghua.edu.cn, }
        \texttt{\small hyangbw@connect.ust.hk,}
        \texttt{\small fuxiaoyi@kongfoo.cn,}
        \texttt{\small qyaoaa@tsinghua.edu.cn}
}

\begin{document}
\maketitle
\begin{abstract}
Heterogeneous Graph Neural Networks (HGNNs) are a class of deep learning models designed specifically for heterogeneous graphs, which are graphs that contain different types of nodes and edges. This paper investigates the application of curriculum learning techniques to improve the performance and robustness of Heterogeneous Graph Neural Networks (GNNs). To better classify the quality of the data, we design a loss-aware training schedule, named LTS that measures the quality of every nodes of the data and incorporate the training dataset into the model in a progressive manner that increases difficulty step by step. LTS can be seamlessly integrated into various frameworks, effectively reducing bias and variance, mitigating the impact of noisy data, and enhancing overall accuracy. Our findings demonstrate the efficacy of curriculum learning in enhancing HGNNs capabilities for analyzing complex graph-structured data. The code is public at \url{https://github.com/LARS-research/CLGNN/}. 
    \footnote{Correspondence is to Q. Yao.}
\end{abstract}

\section{Introduction}

Graphs are indeed a fundamental data structure that accurately represents many real-world relationships and systems, ranging from social and rating networks~\cite{newman2002random} to biology networks~\cite{gilmer2017neural}. In traditional Graph Neural Networks (GNNs), the types of nodes and edges are typically fixed, which imposes certain limitations when dealing with complex relationships and heterogeneous information. To better handle different types of nodes and edges, Heterogeneous Graph Neural Networks (HGNNs) have emerged.

 Despite the numerous HGNN-based node classification approaches that have been proposed, such as those cited in references ~\cite{chen2020simple,kipf2016semi,wu2019simplifying,hamilton2017inductive}, these approaches often assume uniform contribution from all training nodes. However, in reality, the quality of training nodes can vary significantly. Given that HGNNs rely on data-driven methodologies, their performance can suffer when trained on low-quality nodes. 

Given the \textit{ogbn-mag} dataset, which is part of the Open Graph Benchmark~\cite{hu2020open} and represents a heterogeneous network composed of a subset of the Microsoft Academic Graph (MAG) ~\cite{wang2020microsoft}. This dataset comprises four entity types and four types of directed relations linking two entity types. The primary objective revolves around predicting the venue (conference or journal) of each paper, leveraging factors such as content, references, authors, and authors’ affiliations. In the context of the OGB leaderboard, existing methods, such as SeHGNN ~\cite{yang2023simple}, RpHGNN ~\cite{hu2023efficient}, PSHGCN ~\cite{he2023spectral}, etc., predominantly revolve around architecture of HGNNs that capture characteristics of neighboring nodes. Some of these methods also incorporate embedding methods, like LINE ~\cite{tang2015line}, ComplEx ~\cite{trouillon2016complex} and metapath2vec ~\cite{dong2017metapath2vec} into their existing HGNNs frameworks. 

To address the limitation of existing method on OGB leaderboard which do not accurately represent the characteristics of the data, we proposed an innovative curriculum learning method called Loss-aware Training Schedule (LTS) . In our work, we demonstrate how LTS enhances the overall accuracy of different models. With the integration of LTS, our model consistently outperforms the latest GNN models and has secured the top position on the \textit{ogbl-mag} leaderboard for node classification. This highlights the effectiveness of our approach in addressing the limitations of existing CL and optimizing model performance. 

\section{Related Work}
\subsection{GCN}
Graph Neural Networks (GNNs) is a type of neural network specifically designed to work with graph-structured data. For instances, Graph Convolution Network (GCN) \cite{kipf2016semi} has gained significant attention in recent years for its effectiveness in learning representations of graph-structured data.In each layer, the primary operation involves message passing, wherein information is aggregated from the 1-hop neighbors of each vertex. This process contributes to the enhancement of the vertex's semantic representation by incorporating knowledge from its immediate neighbors in the graph. 

Simple and Efficient Heterogeneous Graph Neural Network (SeHGNN) ~\cite{yang2023simple} efficiently handles diverse graph data by simplifying structural information capture through pre-computed neighbor aggregation. This reduces complexity and eliminates redundant calculations. It adopts a single-layer structure with extended metapaths and a transformer-based semantic fusion module for effective feature fusion. SeHGNN's design yields a simple network, high prediction accuracy, and fast training speed.

A hybrid pre-computation-based GNN, Random Projection Heterogeneous Graph Neural Network (RpHGNN) \cite{hu2023efficient}, merges efficiency with low information loss by employing propagate-then-update iterations. It features a Random Projection Squashing step for linear complexity growth and a Relation-wise Neighbor Collection component with an Even-odd Propagation Scheme for finer neighbor information gathering.

\subsection{Curriculum Learning}
Curriculum learning (CL) is a widely recognized approach inspired by educational psychology where the learning process is organized in a way that gradually increases the complexity or difficulty of the tasks presented to the model. This approach mimics the way humans learn as it introduces training data from simple to complex samples. A general CL framework incorporates a difficulty measurer and a training schedule. The CL strategy, as a user-friendly plugin, has shown its efficacy in enhancing the generalization capability and convergence speed of numerous models in a wide range of scenarios such as computer vision and natural language processing and so on ~\cite{wang2021survey}. Nevertheless, accurately assessing the difficulty or quality of nodes within a graph possessing dedicated topology remains an unresolved challenge.

\section{Proposed Methods}
Our framework, LTS aims to evaluate node difficulty or quality within a graph and strategize the training schedule, resulting in enhanced performance for node classification. 
\subsection{Loss-aware Training Schedule}
The topology of a graph is intricate and complex, making it difficult to ensure that various models effectively assess the difficulty and quality of nodes based solely on topological features. To address this challenge, we opt for an approach grounded in machine learning principles, specifically from the perspective of models. We utilize the magnitude of the loss function for each node during training as an indicator of the model's perceived difficulty in learning that node. Firstly, this method aligns more closely with the paradigm of machine learning, eliminating the need for manual adjustments or empirical tuning to adapt the model for curriculum learning. Secondly, it is applicable to any model, requiring only modifications to the loss function without introducing additional computational overhead. To optimize training, nodes are ranked based on their loss function magnitudes, reflecting their learning difficulties. Beginning with nodes exhibiting lower loss values, we progressively introduce those with higher values, ensuring a balanced approach. This prevents undue influence from noisy nodes, thus enhancing model robustness. The algorithm is outlined in Algorithm~\ref{alg:alg}:

  \begin{algorithm}[ht]
    \caption{LTS: Nodes Difficulty Measure and  Strategize Training Schedule}
    \label{alg:alg}
    \begin{algorithmic}[1]

      \STATE \textbf{Calculate loss}: \\
        $loss = loss\_func(reduction='none')$
        \STATE \textbf{Sort loss}: \\
        $indices = sort(loss)$
        \STATE \textbf{Strategize Training Schedule }: \\
        $size = training\_schedule()$ \\
        $num\_large\_losses = len(losses) * size$\\
        $idx = indices[:num\_large\_losses]$\\
        $loss =loss[idx].mean()$\\
        \STATE  \textbf{Loss backward}: \\
        loss.backward()
    \end{algorithmic}
  \end{algorithm}

A training schedule function maps each training epoch $t$ to a
scalar $\lambda_t \in (0, 1]$, , indicating the proportion of the easiest training nodes utilized at the $t$-th epoch. Denoting $\lambda_0$ as the initial proportion of available easiest nodes and $T$ as the epoch when the training schedule function first reaches 1, three pacing functions are considered: linear, root, and geometric.The details of training schedule is defined as in Algorithm~\ref{alg:TS}:
  \begin{algorithm}[ht]
    \caption{Strategize Training Schedule}
    \label{alg:TS}
    \textbf{def}\quad training\_schedule(lam,t,T,scheduler):
    \begin{algorithmic}[1]
    \IF{$scheduler == 'linear'$} \RETURN {$min(1, lam + (1 - lam) * t / T)$} \ELSIF{$scheduler == 'root'$} \RETURN {$min(1, sqrt(lam^2 + (1 - lam^2) * t / T))$}\ELSIF{$scheduler == 'geom'$} 
    \RETURN {$min(1, 2 ^{(log2(lam) - log2(lam) * t / T)})$}\ENDIF
               
    \end{algorithmic}
  \end{algorithm}
\\
Furthermore, training does not halt abruptly at $t=T$, as the backbone GNN may not have thoroughly explored the knowledge of recently introduced nodes at this juncture. Rather, for $t>T$, the entire training set is employed to train until the test accuracy on the validation set reaches convergence.
\section{Experiments}
In this section, we will thoroughly explore and scrutinize the performance of the proposed LTS method. Through a variety of thorough experiments, we have demonstrated the effectiveness of our method by demonstrating increased accuracy in predictions.
\subsection{Experimental Setup}

\subsubsection{Dataset}
    We evaluate our LTS on the Open Graph Benchmark dataset, \textit{ogbn-mag}. 
    The \textit{ogbl-mag} dataset is a heterogeneous, directed graph which contains  four types of entities—papers, authors, institutions, and fields of study—as well as four types of directed relations connecting two types of entities—an author is “affiliated with” an institution, an author “writes” a paper, a paper “cites” a paper, and a paper “has a topic of” a field of study. Each paper is associated with a 128-dimensional word2vec feature vector, while no input node features are associated with other entity types.The objective is to identify the appropriate academic platform (either a conference or a journal) for each research paper, based on its content, citations, the names of the authors, and the institutions they are affiliated with. This task is particularly relevant since many submissions in the Microsoft Academic Graph (MAG) lack or have uncertain venue information due to the inherent inaccuracies in web-based data. The ogbn-mag dataset encompasses a vast array of 349 distinct venues, transforming this challenge into a multiclass classification task with 349 different classes.
    Statistics of the dataset are provided in Table~\ref{tab:dataset}.

\begin{table}[ht]
    \centering
    \caption{Dataset statistics}
    \small
    \begin{tabular}{c|ccc}
    \toprule
         Nodes type & Train & Validation & Test \\ \midrule
         Papers     & 629,571 & 64,879 & 41,939  \\ 
         Authors & 1,134,649   &  --  & -- \\
         Institutions & 8,740 &  -- & --\\
         Fields of study & 59,965 & --&--\\
    \bottomrule
    \end{tabular}
    \label{tab:dataset}
\end{table}

\subsubsection{Evalution Metric}
OGB provides standardized dataset splits and evaluators that allow for easy and reliable comparison of different models in a unified manner. For \textit{ogbl-mag} dataset, the evaluation metric is accuracy of multi-class classification. More specifically, the accuracy was determined by dividing the count of correct predictions across all 349 classes by the total number of predictions made for those classes.

\subsubsection{Settings}
The experiments were conducted using an NVIDIA GeForce RTX 3090. To demonstrate the effectiveness of our approach, we utilized the publicly available model codes provided by OGB along with their corresponding hyperparameters.


\subsection{Accuracy Improvement}
In this section, we demonstrate how our method enhances node classification results. As shown in Table~\ref{tab:accuracy improved}, method with LTS beats those original algorithm. Notably, our method improves RpHGNN (w/ LP+CR+LINE embs) and achieves 20.4\% more performance improvement.
It indicates our curriculum learning method can accurately assessing the difficulty or quality of nodes within a graph possessing dedicated topology and significantly enhance performance of the GNN models. 

         
\begin{table}
    \centering
    \caption{Node Classification Performance of our method and other GNN models on dataset \textit{ogbn-mag}. 
    \textbf{Bold} indicates the best performance. }
    \small
    \setlength\tabcolsep{2 pt}
    \begin{tabular}{c|cc}
    \toprule
        Methods & Valid.  & Test  \\
        \midrule
         GCN & 0.3344 ± 0.0015& 0.3406 ± 0.0019 \\
         GCN (w/ LTS) &0.3428 ± 0.0008 & 0.3501 ± 0.0016 \\ \midrule
         SeHGNN+ComplEx & 0.5917 ± 0.0009 & 0.5719 ± 0.0012 \\
         SeHGNN+ComplEx\\(w/ LTS) & 0.5953 ± 0.0008 & 0.5759 ± 0.013 \\ \midrule
         RpHGNN+LP+CR+LINE & 0.5973 ± 0.0008	& 0.5773 ± 0.0012 \\
         RpHGNN+LP+CR+LINE\\ (w/ LTS) & \textbf{0.8021 ± 0.0020} & \textbf{0.7956 ± 0.0047} \\
         
         \bottomrule
    \end{tabular}
    \label{tab:accuracy improved}
\end{table}

\section{Conclusion}
In this paper, we introduce LTS, a curriculum learning technique tailored to evaluate node difficulty or quality within a graph and strategize the training schedule. LTS identifies suitable training nodes based on loss in each epoch to train GNN. Moreover, this user-friendly plugin method can be applied across various GNN models. To assess LTS's effectiveness, we conduct experiments on real-world benchmark datasets, consistently demonstrating that the method with LTS outperforms all original approaches. Integrating LTS with RpHGNN sets a new standard for node classification performance on the \textit{ogbn-mag} dataset. As part of future research, we aim to further explore the remarkable results achieved by the combination of LTS and RpHGNN, and validate its performance on various datasets through additional experiments.

 
\bibliography{anthology, custom}
\bibliographystyle{acl_natbib}

\end{document}